%% file: main2.tex
\documentclass{article} 
\usepackage{iclr2026_conference,times}

\input{math_commands.tex}

\usepackage{hyperref}
\usepackage{url}

\iclrfinalcopy

\track{Research}

\usepackage{amsmath, amsfonts}
\usepackage{graphicx}
\usepackage{xcolor}
\usepackage{booktabs}
\usepackage{caption}
\usepackage{subcaption}
\usepackage{xspace}

\usepackage{algorithm}
\usepackage{algpseudocode}

\usepackage{xcolor}

\newcommand{\dataset}{DynLMC\xspace}

\usepackage{amsmath}

\title{\dataset: Dynamic Linear Coregionalization for Realistic Synthetic Multivariate Time Series}

\author{Annita Vapsi\thanks{These authors contributed equally}, Penghang Liu\footnotemark[1], Saheed Obitayo, Aakriti, Manoj Cherukumalli, \\ 
\textbf{Prathamesh Patil}, \textbf{Amit Varshney}, \textbf{Nicolas Marchesotti}, \textbf{Elizabeth Fons}, \textbf{Vamsi K. Potluru}, \\ 
\textbf{Manuela Veloso}\thanks{Work done while at JPMorganChase AI Research}~\\
JPMorganChase \\
\texttt{\{penghang.liu, elizabeth.fons, vamsi.k.potluru\}@jpmchase.com} 
}

\begin{document}
\maketitle

\begin{abstract}
Synthetic data is essential for training foundation models for time series (FMTS), but most generators assume static correlations, and are typically missing realistic inter-channel dependencies. We introduce \textbf{\dataset}, a \textbf{Dyn}amic \textbf{L}inear \textbf{M}odel of \textbf{C}oregionalization, that incorporates time-varying, regime-switching correlations and cross-channel lag structures. Our approach produces synthetic multivariate time series with correlation dynamics that closely resemble real data. Fine-tuning three foundational models on \dataset -generated data yields consistent zero-shot forecasting improvements across nine benchmarks. Our results demonstrate that modeling dynamic inter-channel correlations enhances FMTS transferability, highlighting the importance of data-centric pretraining.

\end{abstract}

\section{Introduction}


Foundation models for time series (FMTS) aim for broad forecasting generalization across domains by learning transferable time-series priors. Their goal is to support forecasting, probabilistic prediction, imputation, anomaly detection, and classification with minimal adaptation.
Time series vary widely across domains like healthcare, finance, and weather, ranging from univariate to multivariate signals with nonstationarity and dynamic cross-series correlations, which complicates transfer.
Because real, large-scale multivariate data with realistic dependencies are hard to obtain, synthetic data pretraining offers a scalable path to coverage.
Recent models such as Chronos~\citep{ansari2024chronos} and TimePFN~\citep{timepfn} demonstrate that synthetic priors yield strong zero-shot and few-shot performance, and that the diversity and fidelity of the generator drive transferability.

\begin{figure}
\centering
\includegraphics[width=0.38\textwidth]{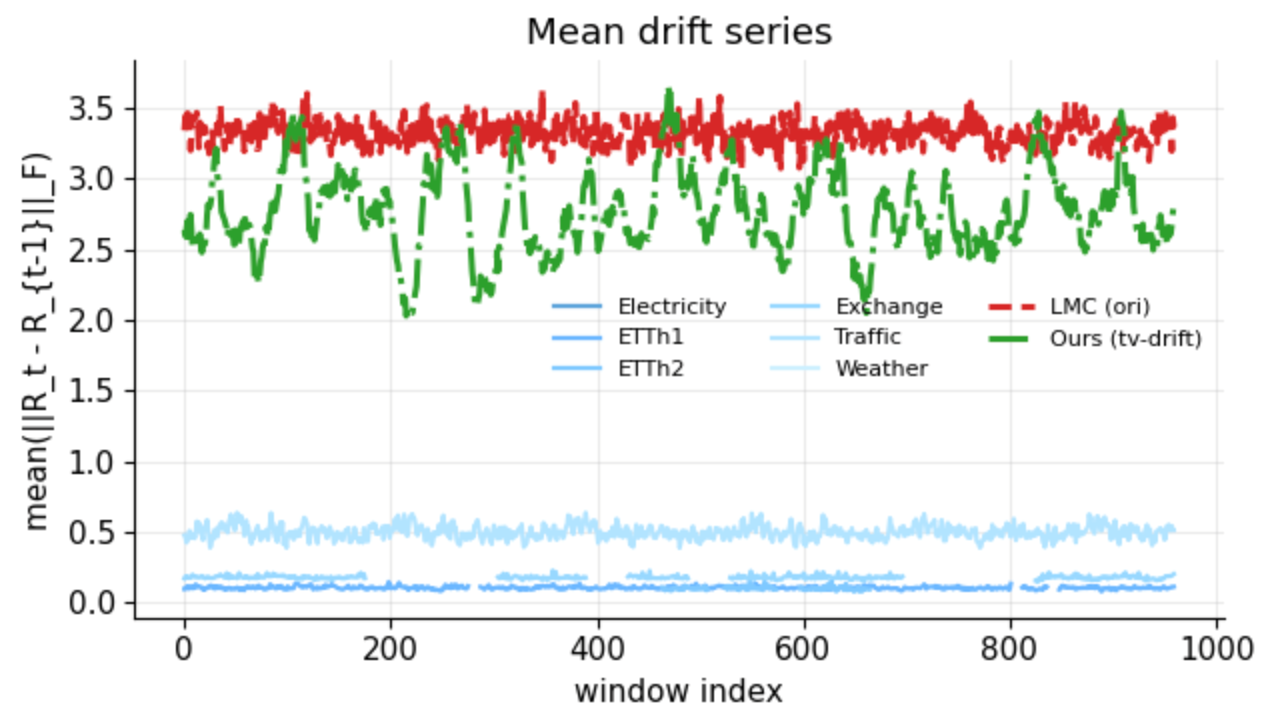}~
\includegraphics[width=0.62\textwidth]{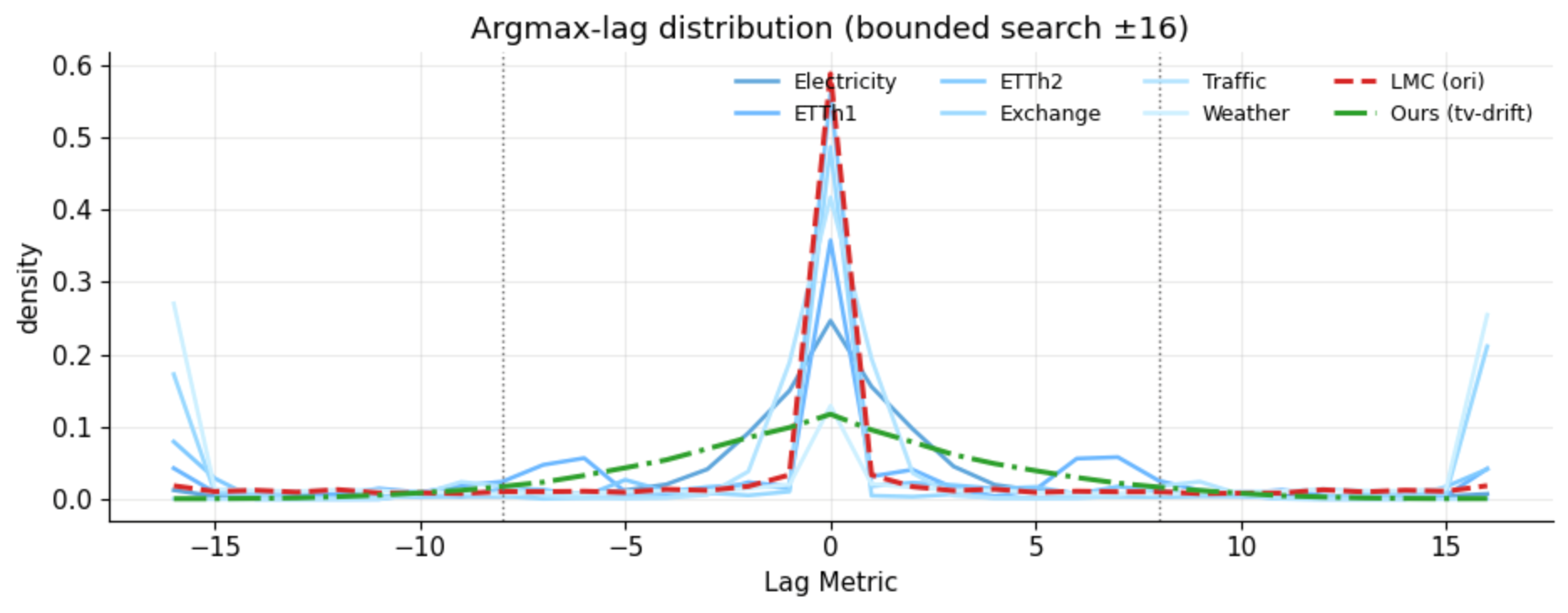}
\caption{\textbf{Left:} Correlation drift comparison across real and synthetic datasets. \textbf{Right:} Argmax-lag distribution comparison across real-world and synthetic datasets.}
\label{fig:driftlag}
\end{figure}

However, current synthetic generators rely on static correlation assumptions. Chronos uses univariate Gaussian process (GP)  sampling, while TimePFN employs the Linear Model of Coregionalization (LMC) for multivariate data, assuming fixed, instantaneous correlations. This produces diverse samples but fails to capture evolving correlations, lagged dependencies, and regime shifts seen in real systems. Our analyses reveal that real datasets exhibit greater temporal variability in cross-channel relationships than LMC-generated data (Figure~\ref{fig:driftlag}). Pretraining on static-correlation data can lead to rigid relational priors, limiting adaptability to real-world signals. 




To address this, we introduce \dataset, which generalizes TimePFN’s generator. \dataset\ extends LMC by: (1) modeling smooth correlation drift via autoregressive updates, (2) introducing regime-switching correlations with a Hidden Markov Model, and (3) incorporating lagged cross-channel dependencies. These mechanisms generate realistic, nonstationary multivariate time series, better reflecting real-world domains like energy, finance, and climate. An illustration of \dataset has been included in Figure~\ref{fig:framework_alt} in the Appendix.

Our main contributions are:
\begin{itemize} 
    \item We present DynLMC, a synthetic dataset generator for multivariate time series that captures evolving and lagged inter-channel dependencies, enabling realistic, nonstationary correlation structures. 
    \item We empirically show that fine-tuning pretrained forecasting models on DynLMC improves robustness and generalization, yielding consistent gains on real-world test datasets. 
\end{itemize}

\section{Background}

\paragraph{Notation and Preliminaries.} Let $\mathcal{D} := \{t, X_t\}_{t=1}^{T}$ denote an $N$-channel multivariate time series of length $T$, where each observation at time $t$ is given by
\[
X_t = [x_{t,1}, x_{t,2}, \ldots, x_{t,N}]^\top \in \mathbb{R}^N,
\]
representing the values of all $N$ channels at that time step.  
Each channel $x_{t,i}$ can depend on its own past values as well as those of other channels, so the entire sequence $X_{1:T} \in \mathbb{R}^{T \times N}$ jointly encodes both temporal and inter-channel relationships.

\paragraph{Static LMC for Synthetic Data.} In the LMC model used by ~\cite{timepfn}, an $N$-variate series $C(t) \in \mathbb{R}^N$ is generated as:
\begin{equation}
C_i(t) = \sum_{j=1}^{L} \alpha_{i,j} \, l_j(t),
\end{equation}
where $l_j(t)$ are independent latent Gaussian processes (GPs) drawn from a diverse kernel bank, and $\alpha_{i,j}$ are fixed coregionalization weights.  
This induces a constant cross-channel covariance:
\[
\text{Cov}(C_i, C_k) = \sum_j \alpha_{i,j}\alpha_{k,j} \text{Cov}(l_j, l_j),
\]
yielding stationary, instantaneous correlations.  

While this framework captures realistic intra-channel variation (trend, seasonality), it cannot model evolving or lagged inter-channel dependencies. In the next section, we extend LMC to dynamically incorporate these structures.

\section{\dataset: Dynamic Linear Model of Coregionalization}

To generate synthetic multivariate time series with evolving and lagged inter-channel dependencies, we introduce the \dataset\ process, which extends the static LMC framework by allowing dynamic mixing weights and random lags between latent and observed channels. This produces datasets with realistic, non-stationary correlation structures, such as correlation drift, regime shifts, and lead-lag effects. The full procedure is detailed in Algorithm~\ref{alg:dynlmc-algo} in the Appendix.

\subsection{Generative Process}

Given $L$ latent GP functions $\{l_j(t)\}_{j=1}^{L}$ sampled via KernelSynth, the \dataset\ process generates each observed channel as:
\begin{equation}
C_i(t) = \sum_{j=1}^{L} \alpha_{i,j}(t) \, l_j(t - \tau_{i,j}),
\label{eq:dynlmc}
\end{equation}
where, $\alpha_{i,j}(t)$ are \textbf{time-varying mixing weights}, modeled either as smooth autoregressive drifts or discrete regime changes, and $\tau_{i,j}$ are integer lags sampled from $[-\tau_{\max}, \tau_{\max}]$, inducing causal offsets between latent and observed channels.

\subsection{Dynamic Correlation Mechanisms}
\label{sec:dyn_correlation_mechanisms}

\paragraph{(a) Smooth Drift.}
Weights evolve as an autoregressive process:
\begin{equation}
\tilde{\alpha}_{i,j}(t) = \rho\, \tilde{\alpha}_{i,j}(t-1) + \epsilon_{i,j}(t), \quad 
\epsilon_{i,j}(t) \sim \mathcal{N}(0, \sigma^2),
\end{equation}
with $\rho \in (0.9, 0.99)$ controlling correlation persistence and are normalized by a softmax at each $t$:
\[
\alpha_{i,j}(t) = \frac{\exp(\tilde{\alpha}_{i,j}(t))}{\sum_k \exp(\tilde{\alpha}_{i,k}(t))}.
\]
Larger $\rho$ yields slower, smoother drift (stronger temporal persistence), while smaller $\rho$ produces faster-changing weights.

\paragraph{(b) Regime Switching (HMM).}
We define $K$ latent regimes $\{s_t\}$ with transition probabilities $p(s_t|s_{t-1})$.  
Each regime $k$ has a fixed mean matrix $\bar{\alpha}^{(k)}$.  
At each time step:
\[
\alpha_{i,j}(t) = (1-\eta)\bar{\alpha}^{(s_t)}_{i,j} + \eta \epsilon_{i,j}(t),
\]
where $\eta$ controls intra-regime variability, introducing structured, interpretable correlation regimes.

\paragraph{(c) Lagged Dependencies.}
To simulate lead–lag effects, we sample delays $\tau_{i,j} \sim \mathcal{U}[-\tau_{\max}, \tau_{\max}]$.  
Positive $\tau_{i,j}$ means latent $j$ leads observed $i$.
A larger $\tau_{\max}$ gives a larger temporal offset between the leading and lagging signals.

\section{Experiments}


We assess the effectiveness of the synthetic data by fine-tuning a pretrained multivariate time series forecasting model on \dataset-generated data and evaluating robustness and generalization on real data. Details of the fine-tuning strategy are provided in section~\ref{ftstrategy} of the Appendix.

\paragraph{Models.}
We evaluate three multivariate time series architectures, TimePFN~\citep{timepfn}, iTransformer~\citep{itransformer}, and Chronos-2~\citep{chronos2}, representing distinct modeling paradigms to assess the impact of our generated data.

\paragraph{Datasets.}
Our dataset suite comprises three primary datasets (Section~\ref{sec:dyn_correlation_mechanisms}): (i) \emph{\dataset\ - Drift} ($\rho$ varies from 0.92 to 0.99), (ii) \emph{\dataset\ - Lag} ($\tau_{\max}$ varies from 0 to 8), and (iii) \emph{\dataset\ - Regime Shift} ($K$ varies from 2 to 6). We also include two combined variants: (iv) \emph{\dataset\ - Combined}, which mixes equal weights of each primary dataset, and (v) \emph{\dataset\ - Combined BO}, which uses Bayesian optimization to determine optimal mixing weights (see Section~\ref{sec:bo_mixing}). Each synthetic dataset contains 1,500 samples, 160 channels, and 1,024 time steps. For the Bayesian optimized variant, the mixing algorithm determines each primary dataset contribution.

\paragraph{Training protocol.}
For TimePFN and iTransformer, we pretrain each model using the original  LMCSynth synthetic data, dataset from TimePFN, then fine-tune the pretrained checkpoints on each of the four \dataset datasets. For Chronos-2, we use the official pretrained checkpoint due to unavailable pretraining data and similarly fine-tune on each synthetic dataset. Further details on fine-tuning and training infrastructure are in section~\ref{ftstrategy} of the Appendix.

\paragraph{Testing protocol.} 
We evaluate both pretrained and fine-tuned models on nine real benchmark datasets: ECL, ETTh1/2, ETTm1/2, Exchange, Solar, Traffic, and Weather. Dataset descriptions are in Appendix section~\ref{testingdatasets}.

\begin{table*}[h!]
\centering
\caption{Mean Absolute Error (MAE) performance and number of wins over pretrained baseline, across benchmarks (averaged over 5 seeds). KS = KernelSynth, LMC = static LMCSynth generator, and DynLMC = our dynamic extension with time-varying correlations and lag-structures. The best result is highlighted in bold and the second-best is underlined.}
\label{tab:main_results}
\small
\setlength{\tabcolsep}{5pt}
\resizebox{\textwidth}{!}{
\begin{tabular}{lccccccccc|cc}
\toprule
\textbf{Model} & \textbf{ECL} & \textbf{ETTh1} & \textbf{ETTh2} & \textbf{ETTm1} & \textbf{ETTm2} & \textbf{Exchange} & \textbf{Solar} & \textbf{Traffic} & \textbf{Weather} & \textbf{Wins} \\
\midrule
\multicolumn{11}{l}{\textbf{TimePFN Backbone}} \\
\midrule
Baseline pretrained on KS + LMC & 0.369 & 0.440 & \underline{0.369} & \underline{0.483} & \underline{0.298} & \underline{0.226} & 1.063 & \underline{0.596} & \underline{0.297} &  -- \\
Fine-tuned (KS + LMC + \dataset\ - Drift) & 0.371 & \textbf{0.436} & 0.374 & 0.496 & 0.299 & \textbf{0.225} & 1.077 & \underline{0.596} & \underline{0.297} & 2 \\
Fine-tuned (KS + LMC + \dataset\ - Lag) & 0.456 & 0.459 & 0.382 & 0.529 & 0.321 & 0.229 & \textbf{0.967} & 0.666 & 0.300 & 1\\
Fine-tuned (KS + LMC + \dataset\ - Regime Shift) & \underline{0.360} & 0.438 & \underline{0.369} & 0.492 & 0.301 & 0.229 & \underline{1.009} & \textbf{0.593} & 0.301 & 4 \\
Fine-tuned (KS + LMC + \dataset\ - Combined) & \textbf{0.356} & \underline{0.437} & \textbf{0.367} & \textbf{0.476} & \textbf{0.297} & 0.228 & 1.021 & \underline{0.596} & \textbf{0.294} & 7 \\
\midrule
\multicolumn{11}{l}{\textbf{iTransformer Backbone}} \\
\midrule
Baseline pretrained on KS + LMC & 0.472 & 0.462 & 0.367 & 0.563 & 0.307 & \underline{0.226} & 0.844 & 0.676 & \underline{0.279} & --  \\
Fine-tuned (KS + LMC + \dataset\ - Drift) & 0.456 & 0.468 & 0.369 & 0.518 & 0.295 & 0.233 & 0.763 & 0.638 & \textbf{0.278} & 7  \\
Fine-tuned (KS + LMC + \dataset\ - Lag) & 0.436 & 0.458 & 0.368 & 0.524 & 0.303 & 0.232 & \textbf{0.735} & 0.672 & \underline{0.279} & 7 \\
Fine-tuned (KS + LMC + \dataset\ - Regime Shift) & \underline{0.376} & \underline{0.448} & \textbf{0.358} & \underline{0.513} & \underline{0.294} & \underline{0.226} & 0.781 & \underline{0.610} & \textbf{0.278} & 9 \\
Fine-tuned (KS + LMC + \dataset\ - Combined) & \textbf{0.368} & \textbf{0.447} & \underline{0.364} & \textbf{0.458} & \textbf{0.288} & \textbf{0.225} & \underline{0.762} & \textbf{0.603} & 0.286 & 8 \\
\midrule
\multicolumn{11}{l}{\textbf{Chronos-2 Backbone}} \\
\midrule
Baseline (pretrained checkpoint) & \textbf{0.273} & \textbf{0.397} & 0.344 & 0.452 & 0.273 & \textbf{0.209} & \textbf{0.343} & \textbf{0.309} & \textbf{0.230} & --  \\
Fine-tuned (KS + LMC + \dataset\ - Drift) & 0.286 & \underline{0.399} & 0.345 & \textbf{0.400} & \textbf{0.264} & 0.217 & 0.370 & 0.337 & 0.250 & 2  \\
Fine-tuned (KS + LMC + \dataset\ - Lag) & 0.290 & 0.401 & 0.348 & \underline{0.406} & \underline{0.267} & \underline{0.216} & \underline{0.348} & 0.333 & \underline{0.243} & 2 \\
Fine-tuned (KS + LMC + \dataset\ - Regime Shift) & 0.286 & 0.406 & \textbf{0.336} & 0.418 & 0.268 & 0.228 & 0.368 & \underline{0.330} & 0.251 & 3 \\
Fine-tuned (KS + LMC + \dataset\ - Combined) & \underline{0.285} & 0.406 & 0.343 & 0.422 & 0.274 & 0.231 & 0.417 & 0.354 & 0.248 & 2 \\
\midrule
\multicolumn{11}{l}{\textbf{Naïve Predictions}} \\
\midrule
Naïve Mean & 0.794 & 0.558 & 0.387 & 0.548 & 0.307 & 0.269 & 0.734 & 0.805 & 0.271 & --  \\
Naïve Last & 0.996 & 0.713 & 0.422 & 0.665 & 0.328 & \textbf{0.196} & 0.815 & 1.077 & \textbf{0.254} & --  \\
Seasonal Naïve & 1.017 & 0.727 & 0.431 & 0.674 & 0.334 & 0.205 & 0.844 & 1.098 & 0.263 & -- \\
\bottomrule
\end{tabular}
}
\end{table*}



\subsection{Zero-Shot Forecasting Results}
Table~\ref{tab:main_results} shows the mean absolute error (MAE) for pretrained TimePFN, iTransformer, and Chronos-2 models, as well as their performance after fine-tuning on each \dataset\ variant. Results for the \dataset\ - Combined BO variant (TimePFN only) are in Table~\ref{tab:bo_results} in the Appendix. TimePFN and iTransformer are pretrained solely on synthetic data, while the public Chronos-2 model was also pretrained on real data.

\paragraph{Performance on TimePFN and iTransformer.} 
Fine-tuning on \dataset\ consistently improves TimePFN and iTransformer performance, especially on highly multivariate datasets like ECL, Traffic, and Weather. For TimePFN, the Regime Shift and Combined variants achieve 4 and 7 wins, respectively, over the baseline. For iTransformer, Regime Shift and Combined yield 9 and 8 wins, showing that modeling regime shifts and mixed dependencies during synthetic pretraining significantly enhances generalization and robustness.

\paragraph{Performance on Chronos-2.} 
For Chronos-2, improvements are more modest. The Regime Shift variant achieves 3 wins over the official checkpoint, while other variants have smaller or mixed effects. This aligns with the Chronos-2 training, which already includes synthetic data with evolving inter-channel dependencies, reducing the marginal benefit of additional synthetic fine-tuning with \dataset. Additionally, the Chronos-2 pretraining on both synthetic and real data likely contributes to its stronger baseline and limits further gains from synthetic fine-tuning.

\section{Conclusion}

We presented \textbf{\dataset}, a dynamic extension of the Linear Model of Coregionalization for realistic synthetic multivariate time series.  
By incorporating smooth drift, regime-switching, and lagged dependencies, \dataset\ bridges the realism gap between synthetic and real data.  
Fine-tuning TimePFN with \dataset\ improves zero-shot forecasting without modifying architecture or training objective.  
This work highlights the role of realistic inter-channel dynamics as a key ingredient in synthetic pretraining for time-series foundation models.

\section{Disclaimer}
This paper was prepared for informational purposes  by the Artificial Intelligence Research group of JPMorgan Chase \& Co. and its affiliates (''JP Morgan'') and is not a product of the Research Department of JP Morgan. JP Morgan makes no representation and warranty whatsoever and disclaims all liability, for the completeness, accuracy or reliability of the information contained herein. This document is not intended as investment research or investment advice, or a recommendation, offer or solicitation for the purchase or sale of any security, financial instrument, financial product or service, or to be used in any way for evaluating the merits of participating in any transaction, and shall not constitute a solicitation under any jurisdiction or to any person, if such solicitation under such jurisdiction or to such person would be unlawful.

\bibliography{aaai2026}

\bibliographystyle{iclr2026_conference}


\appendix
\section{Appendix}

\subsection{Related Work}
\paragraph{Time Series Foundation Models.} Recent work has investigated foundation models for time series, aiming to pretrain large neural architectures on diverse temporal data and transfer them across tasks and domains. TimesFM~\cite{timesfm} and Chronos~\cite{ansari2024chronos} pioneered this direction by framing forecasting as a sequence modeling problem and training transformers on large-scale synthetic univariate time series. The Chronos~\cite{ansari2024chronos} synthetic generation pipeline, KernelSynth, captures common temporal structures such as trends, seasonality, and stochastic noise, enabling strong zero-shot and few-shot performance across heterogeneous benchmarks without reliance on curated real-world datasets. These approaches predominantly focus on univariate series or treat multiple variables independently, limiting their capacity to model cross-variable dependencies inherent in multivariate data.

\paragraph{Multivariate Time Series Modeling.} Modeling dependencies across variables is central to multivariate time series forecasting. Classical approaches rely on vector autoregressive or state-space models, while modern deep learning methods employ attention mechanisms and graph-based inductive biases. iTransformer~\cite{itransformer} introduced an inverted attention mechanism that attends across variables rather than time steps, explicitly targeting multivariate dependency modeling and achieving strong empirical performance.

In parallel, models such as Toto~\cite{toto}, a general-purpose time series forecasting model specifically tuned for observability metrics, and TTMs~\cite{ttms}, a compact model based on the TSMixer architecture, explored alternative pretraining objectives, tokenization strategies, and transformer architectures for time series foundation models.

Models such as Moirai~\cite{moirai} and Moirai-2~\cite{moirai2} extend foundation-model-style training to multivariate settings, emphasizing scalability and robustness across datasets with varying dimensionality. However, these methods primarily rely on real-world multivariate datasets or implicit modeling of inter-variable structure, offering limited control over the statistical properties of cross-series relationships during pretraining.

More recent work in Chronos-2~\cite{chronos2} refined architectural choices and scaling behavior, including support for univariate, multivariate, and covariate-informed forecasting and introducing a new group attention mechanism sharing information across multiple time series within a group.

\paragraph{Synthetic Data for Multivariate Pretraining.} Synthetic data generation has proven critical for scaling time series foundation models. Chronos~\cite{ansari2024chronos} demonstrated that synthetic pretraining can induce useful inductive biases and broad generalization. However, its generation pipeline assumes independent univariate series and does not model interactions between variables.
TimePFN bridges this gap by extending the Chronos synthetic generation framework to multivariate time series through correlated covariates, enabling the model to learn cross-variable dependencies during pretraining. This approach shows improved performance on multivariate downstream tasks, particularly in data-scarce regimes. 

Nevertheless, the generated dependencies are largely static, assuming fixed correlation structures across time, and often across all variates. In contrast, real-world multivariate time series frequently exhibit non-stationary dependency patterns, including correlation drift, lagged interactions, and regime-dependent relationships between variables. Such dynamics are common in financial, environmental, and sensor data but remain underrepresented in existing synthetic pretraining pipelines. Concurrent work to ours in Chronos-2~\cite{chronos2} introduced the concept of multivariatizers, sampling multiple time series from base univariate generators and introducing correlations between variates in a series and dependencies across time. 

Our work extends synthetic multivariate pretraining by explicitly modeling dynamic inter-variable structure. Building on Chronos~\cite{ansari2024chronos} and TimePFN~\cite{timepfn}, we introduce synthetic generators that incorporate correlation drift, variable-specific lags, and regime shifts between covariates. By enriching the synthetic training distribution with these realistic dependency dynamics, we enable multivariate foundation models to better generalize to non-stationary and structurally complex real-world datasets. While our contributions in introducing inter-dependencies between variates overlap with Chronos-2~\cite{chronos2}, we are making our multivariate time series generation pipeline available upon request and show that fine-tuning the open-source Chronos-2~\cite{chronos2} model with the \dataset\ data improves its performance in several real datasets.

\newpage

\subsection{\dataset Illustration}

\begin{figure}[H]
    \centering
    \includegraphics[width=0.95\linewidth]{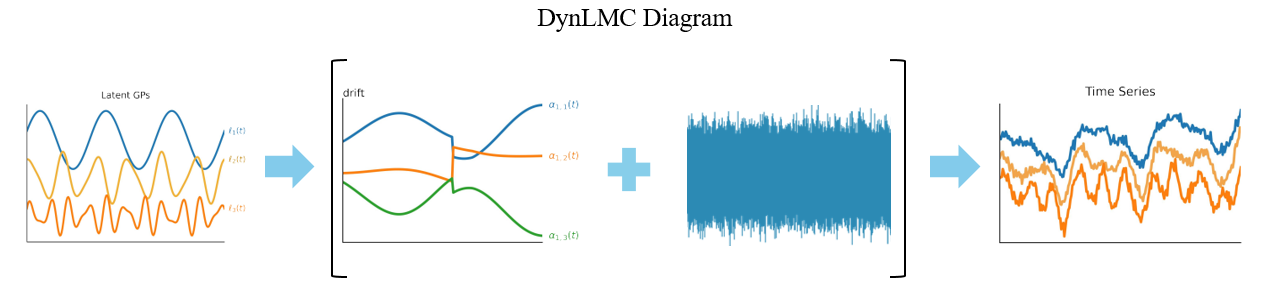}
    \caption{An illustration of \dataset\ with evolving inter-channel structures (drift).}
    \label{fig:framework_alt}
\end{figure}

\newpage

\subsection{\dataset\ Dataset Generation Algorithm}

\begin{algorithm}[]
\caption{\dataset\ Generation Algorithm}
\label{alg:dynlmc-algo}
\textbf{Input:} Number of variates $N$, time-series length $T$, Weibull shape $\beta$, Weibull scale $\lambda$, 
minimum number of latent functions $m$, maximum number of kernel compositions in KernelSynth $J$, 
drift parameters $(\rho,\sigma)$, regime type $r \in \{\texttt{none},\texttt{hmm}\}$, 
number of HMM regimes $K$, HMM stickiness $\kappa$, lag bound $\tau_{\max}$\\
\textbf{Output:} Synthetic MTS $C$ with $N$ variates and length $T$
\begin{algorithmic}[1]
\State $L \sim \max(\min(\mathrm{Weibull}(\beta,\lambda), N), m)$
\For{$j \in \{1 \ldots L\}$}
    \State $l_j(t) \leftarrow \mathrm{KernelSynth}(J, T)$
\EndFor
\State $\tau_{i,j} \sim \mathrm{Unif}\{-\tau_{\max},\ldots,\tau_{\max}\}$ for all $(i,j)$
\If{$r = \texttt{none}$}
    \State For all $(i,j)$: define logits $A_{i,j}(t) \leftarrow \mathrm{AR1Path}(T;\rho,\sigma)$
\Else
    \State Sample HMM states $s_{1:T}$ with stickiness $\kappa$ and regime means $\bar{\alpha}^{(k)}$
    \State For all $(i,j)$: $A_{i,j}(t) \leftarrow \bar{\alpha}^{(s_t)}_{i,j} + \mathrm{AR1Path}(T;\rho,\sigma)$
\EndIf
\State For each $t$: $\big[\alpha_{i,1}(t),\ldots,\alpha_{i,L}(t)\big] \leftarrow \mathrm{softmax}\!\big(A_{i,1:L}(t)\big)$ for all $i$
\For{$i \in \{1 \ldots N\}$}
    \State $C_i(t) \leftarrow \sum_{j=1}^{L} \alpha_{i,j}(t)\, l_j\!\big(t-\tau_{i,j}\big)$ \hfill for $t=1\ldots T$
\EndFor
\State \textbf{return} $\{C_i(t)\}_{i=1}^{N}$
\end{algorithmic}
\end{algorithm}

\newpage

\subsection{Bayesian Optimization for Synthetic Dataset Mixing}
\label{sec:bo_mixing}
Our dataset generation model \dataset\ provides methods for generating the three distinct types of multivariate time series data, as described in Section~\ref{sec:dyn_correlation_mechanisms}: (i) \dataset\- Drift, where inter-variable correlations evolve smoothly over time; (ii) \dataset\- Lag, where dependencies manifest with temporal offsets between variates; and (iii) \dataset\- Regime Shift, characterized by abrupt changes in the underlying correlation structure. Finally, we combine the three datasets with equal weights to create a fourth variant (iv) \dataset\ - Combined, allowing us to assess whether incorporating all three datasets into the training process yields any additional performance improvements. 

Beyond this naive mixing we also investigate the potential of a data centric generative optimization procedure, (v) \dataset\ - Combined BO, that learns an optimal mixture over multiple synthetic time series sources with the objective to optimize the training data distribution for downstream forecasting accuracy. This is a time series extension of the optimization framework introduced for tabular data~\cite{hamad2023supervised}. Let $M$ denote the index set of all available synthetic time series sources where each source corresponds to a distinct temporal generative process (e.g drift/lag, regime shift generators) and let $\{S_m\}$ denote the dataset corresponding to the source ${m \in M}$. At each iteration, we construct a synthetic training mixture by sampling sliding window sequences ($X, Y$) where $X \in \mathbb{R}^{T \times N}$ and $Y \in \mathbb{R}^{T_{\text{pred}} \times N}$ from each source according to mixture weights $\boldsymbol{\gamma}^{(0)} \in \Delta^{|M|}$. In some occasions where synthetic sources could have different channel dimensionalities, we can  align these windows to a common channel dimensionality $\boldsymbol{N}^\star$ which can be used to train a forecasting model $f_\phi$(in our implementation, a chronos/TimePFN-style transformer encoder. The trained model is then evaluated on a real validation dataset $\mathcal{D}_{\text{val}}$ using a downstream forecasting metric such as mean absolute error (MAE). The resulting validation loss define a black-box objective over a mixture of weights. Concretely, a single evaluation of this objective consist of training the forecasting model on a candidate synthetic mixture and computing its MAE on the validation set:
\begin{equation}
\ell{(\gamma)} = \mathcal{L}(f_\phi(D_{\text{syn}} (\gamma)), D_{\text{val}}).
\label{eq:optim}
\end{equation}
where $\mathcal{D}_{\text{syn}}$ denotes the synthetic dataset sampled from the mixture defined by $\boldsymbol{\gamma}$.
We optimize this objective using sequential based optimization via the Tree-structured Parzen Estimator (TPE) ~\cite{bergstra2011algorithms}, which iteratively proposes mixture weights, retrains the forecasting model on the resulting synthetic dataset, and updates its surrogate model based on observed validation loss. After $K$ trials, we select $\boldsymbol{\gamma}^\star = \arg\min_k \ell^{(k)}$  and get the mixture dataset ($X, Y$) sampled according to $\boldsymbol{\gamma}^\star$. Unlike prior work that applies Bayesian optimization primarily for hyperparameter tuning of time series models ~\cite{zulfiqar2022hyperparameter} or ensemble weighting \cite{du2022bayesian}, our approach treats the training data as the optimization variable. This yields a data centric optimization framework in which the optimal synthetic mixture is defined by its downstream predictive utility on real time series forecasting tasks. Algorithm~\ref{alg:ts-goat}, details this method.

\begin{algorithm}[t]
\caption{Bayesian Optimization for Time Series Dataset Mixing}
\label{alg:ts-goat}
\textbf{Input:} Time-series source generators $\{S_m\}_{m \in M}$;
validation dataset $\mathcal{D}_{\text{val}}$;
Forecasting model $f_\phi$;

Window parameters ($L_{\text{seq}}, L_{\text{label}}, L_{\text{pred}}$);
time split $(\text{holdout$_{frac}$}, \text{holdout$_{tail}$})$;
Number of variates  $N$;
Bayesian optimizer $\mathcal{B}$\\
\textbf{Output:} Optimal mixture weights $\boldsymbol{\gamma}^\star$; 
Mixture dataset $(X^\star, Y^\star)$
\begin{algorithmic}[1]
\State Initialize mixture weights $\boldsymbol{\gamma}^{(0)} \in \Delta^{M}$\;
\State Set channel target $C^\star = \max_m C_m$ 
\For{$k \in \{1 \ldots K\}$}
    \State Compute allocation $n_m^{(k)}=\lfloor \gamma_m^{(k)}N\rfloor$ with $\sum_m n_m^{(k)}=N$\;
    \For{$m \in \{1 \ldots M\}$}
        \State Sample windows $(X_m^{(k)},Y_m^{(k)}) \leftarrow \textsc{SampleWindows}(S_m,n_m^{(k)},L_{\text{seq}},L_{\text{pred}})$\;
        \State Apply channel alignment $X_m^{(k)} \leftarrow \textsc{Align}(X_m^{(k)},C^\star)$ and $Y_m^{(k)} \leftarrow \textsc{Align}(Y_m^{(k)},C^\star)$\;
    \EndFor
    \State Compose mixture dataset $(X^{(k)},Y^{(k)}) = \bigcup_{m\in M} (X_m^{(k)},Y_m^{(k)})$\;
    \State \textbf{Train model with AMP + early stopping:}
    \State \[ 
    \phi^{(k)} \leftarrow \text{Train}(f_\phi,(X^{(k)},Y^{(k)}),\phi_0)
    \]
    \State Subsample validation windows $(X_{\text{val}},Y_{\text{val}}) \leftarrow \textsc{Subsample}(D_{\text{val}},N_{\text{val}})$ and align channels\;
    \State Predict $\hat{Y}_{\text{val}}^{(k)} = f_{\phi^{(k)}}(X_{\text{val}})$\;
    \State Compute loss $\ell^{(k)} = \mathcal{L}(\hat{Y}_{\text{val}}^{(k)},Y_{\text{val}})$\;

    \State \textbf{Update mixture using BO:}
    \State \[
    \boldsymbol{\gamma}^{(k+1)} \leftarrow \mathcal{B}\big(\{(\boldsymbol{\gamma}^{(i)},\ell^{(i)})\}_{i=1}^{k}\big)
    \]
\EndFor
\State Select $k^\star = \arg\min_k \ell^{(k)}$ and set $\boldsymbol{\gamma}^\star=\boldsymbol{\gamma}^{(k^\star)}$\;
\State \textbf{return} $\boldsymbol{\gamma}^\star, (X^\star, Y^\star)$
\end{algorithmic}
\end{algorithm}

\paragraph{Bayesian Optimization Mixing - Results.} Table~\ref{tab:bo_results} compares the mean absolute error (MAE) of the fine-tuned model on the naively combined dataset variant and the fine-tuned model on the mixed dataset variant generated using the Bayesian optimization algorithm from Section~\ref{sec:bo_mixing}.

\begin{table*}[h!]
\centering
\caption{Mean Absolute Error (MAE) performance across datasets combining drift, lag and regime shift characteristics, naively and using a Bayesian mixing method. 
Results averaged over 5 seeds. The best result is highlighted in bold.}
\label{tab:bo_results}
\small
\setlength{\tabcolsep}{5pt}
\resizebox{\textwidth}{!}{
\begin{tabular}{lccccccccc|cc}
\toprule
\textbf{Model} & \textbf{ECL} & \textbf{ETTh1} & \textbf{ETTh2} & \textbf{ETTm1} & \textbf{ETTm2} & \textbf{Exchange} & \textbf{Solar} & \textbf{Traffic} & \textbf{Weather} & \textbf{Wins} \\
\midrule
\multicolumn{11}{l}{\textbf{TimePFN Backbone}} \\
\midrule
Fine-tuned (KernelSynth + LMCSynth + \dataset\ - Combined) & \textbf{0.356} & \textbf{0.437} & 0.367 & \textbf{0.476} & 0.297 & 0.228 & 1.021 & 0.596 & 0.294 & - \\
Fine-tuned (KernelSynth + LMCSynth + \dataset\ - Combined BO) & 0.378 & 0.440 & \textbf{0.363} & 0.502 & \textbf{0.294} & \textbf{0.226} & \textbf{0.882} & \textbf{0.592} & \textbf{0.270} & 6 \\
\bottomrule
\end{tabular}
}
\end{table*}

\newpage

\subsection{Fine-Tuning Strategy}
\label{ftstrategy}

Our objective is to enrich the pretrained model with additional temporal dependency patterns, while avoiding overfitting to any single synthetic distribution. The original \textsc{TimePFN} work adopted a curriculum learning paradigm for multivariate adaptation. Concretely, they warm-started training by fully fine-tuning on KernelSynth, a dataset derived from the Chronos~\cite{ansari2024chronos} univariate generation pipeline and adapted to the multivariate setting by concatenating independent time series into tensors of shape $(N \times T)$, where $N$ denotes the number of channels and $T$ the sequence length. Subsequently, they introduced LMCSynth, a dataset of dependent multivariate time series generated via a linear model of coregionalization (LMC), thereby switching from a regime of no inter-variable dependencies to one including them.

In our setting, however, curriculum training did not yield optimal results. Instead, we observed improved generalization by jointly mixing the data used during pretraining with newly generated synthetic data. Specifically, we interleave samples from KernelSynth, LMCSynth, and our proposed dataset \dataset, which explicitly models dynamic correlation structures. Importantly, the contribution of \dataset\ increases over the course of training. This progressive reweighting encourages the model to first stabilize on simpler or previously seen distributions before gradually incorporating more complex and dynamically evolving dependency patterns.

\paragraph{Progressive Dataset Mixing.}

Let $\mathcal{D}_1$, $\mathcal{D}_2$, and $\mathcal{D}_3$ denote KernelSynth, LMCSynth, and \dataset, respectively. At training index $i$, we define a mixing probability
\begin{equation}
p_{\text{mix}}(i) = \log\!\left(1 + \frac{2i}{M}\right),
\end{equation}
where $M$ is the total dataset size (minimum length across datasets). Since $p_{\text{mix}}(i)$ is monotonically increasing in $i$, the probability of constructing mixed samples grows over time. Mixed samples are formed by concatenating variable subsets from each dataset along the channel dimension and randomly permuting channels to prevent positional bias. Otherwise, with probability $1 - p_{\text{mix}}(i)$, a single dataset is sampled uniformly at random. This design ensures that dynamically correlated samples from \dataset\ are incorporated more frequently in later training iterations, effectively implementing a soft curriculum through probabilistic mixing rather than staged fine-tuning. Algorithm~\ref{alg:dataset_mixing}, which details this method, is included in the appendix.

\begin{algorithm}[t]
\caption{Progressive Interleaved Mixing of Three Datasets}
\label{alg:dataset_mixing}
\begin{algorithmic}[1]
\Require Datasets $\mathcal{D}_1$ (KernelSynth), $\mathcal{D}_2$ (LMCSynth), $\mathcal{D}_3$ (\dataset)
\State $N \leftarrow \min(|\mathcal{D}_1|, |\mathcal{D}_2|, |\mathcal{D}_3|)$
\For{index $i = 0, \dots, N-1$}
    \State $p_{\text{mix}} \leftarrow \log\!\left(1 + \frac{2i}{M}\right)$
    \State Sample $u \sim \text{Uniform}(0,1)$
    \If{$u < p_{\text{mix}}$}
        \State Sample $(x_1, y_1) \sim \mathcal{D}_1[i]$
        \State Sample $(x_2, y_2) \sim \mathcal{D}_2[i]$
        \State Sample $(x_3, y_3) \sim \mathcal{D}_3[i]$
        \State Randomly choose channel split sizes
        \State Concatenate channel subsets:
        \[
        x \leftarrow \text{concat}(x_1, x_2, x_3), \quad
        y \leftarrow \text{concat}(y_1, y_2, y_3)
        \]
        \State Randomly permute channel order
    \Else
        \State Randomly choose $k \in \{1,2,3\}$
        \State Sample $(x, y) \sim \mathcal{D}_k[i]$
    \EndIf
    \State \Return $(x,y)$
\EndFor
\end{algorithmic}
\end{algorithm}

\paragraph{Retention–Adaptation Tradeoff in Chronos-2.} Unlike TimePFN and iTransformer, where we control the full synthetic pretraining pipeline, we do not have access to the original Chronos-2 training corpus. Consequently, during \dataset\ fine-tuning we cannot explicitly preserve the data distribution learned during pretraining.
To mitigate potential overfitting to \dataset, we freeze all but the final three layers and adopt a reduced learning rate. Nevertheless, the absence of the original pretraining distribution may limit our ability to balance knowledge retention and adaptation. This could explain why \dataset\ fine-tuning occasionally degrades performance on certain real-world datasets for Chronos-2, despite producing gains elsewhere.
Importantly, because we compare the official pretrained checkpoint to our \dataset\ fine-tuned version under identical evaluation conditions, the reported differences isolate the contribution of \dataset\ fine-tuning itself.

\newpage

\subsection{Training Infrastructure}
\label{traininfra}
 All experiments were run on an AWS g6e.2xlarge instance with a single NVIDIA L40S GPU (48GB VRAM), 8 vCPUs, and 64GB RAM for both training and evaluation. Fine-tuning samples were created by slicing synthetic data with a 96-step context window and 96-step forecasting horizon. Early stopping with a patience of 200 iterations was used and results are averaged over five random seeds for both pretraining and fine-tuning.

\newpage

\subsection{Testing Datasets}
\label{testingdatasets}
We evaluated all pretrained and fine-tuned models on nine real-world datasets: \emph{ECL}, \emph{ETTh1}, \emph{ETTh2}, \emph{ETTm1}, \emph{ETTm2}, \emph{Exchange}, \emph{Solar}, \emph{Traffic}, and \emph{Weather}. These datasets encompass diverse real-world distributions and capture a range of temporal patterns. Brief descriptions of each dataset are provided below:

\begin{itemize}
    \item ECL: The ECL (Electricity Load Diagrams;~\cite{electricityloaddiagrams20112014_321}) contains hourly electricity consumption data for 321 clients. The dataset includes 18317/2633/5261 samples for train/validation/test splits.
    \item ETTh Family: The ETTH (Electricity Transformer Temperature;~\cite{zhou2021etth}) datasets each contain seven variates. ETTh1 and ETTh2 are sampled hourly, while ETTm1 and ETTm2 are sampled every 15 minutes. ETTh datasets include 8545/2881/2881 samples for train/validation/test splits, and ETTm datasets contain 34465/11521/11521 samples, respectively.
    \item Exchange: The Exchange~\cite{autoformer} dataset offers daily exchange rates for eight countries, with eight variables in total. The dataset includes 5120/665/1422 samples for train/validation/test splits.
    \item Solar: The Solar~\cite{Lai2017ModelingLA} dataset contains power production data from 137 solar power plants, sampled every 10 minutes, yielding 137 variables. The dataset includes 36601/5161/10417 samples for train/validation/test splits.
    \item Traffic: The Traffic~\cite{autoformer} dataset comprises hourly road occupancy rates from 862 locations, yielding 862 variables. The dataset includes 12185/1757/3509 samples for train/validation/test splits.
    \item Weather: The Weather~\cite{autoformer} dataset consists of 21 meteorological variables collected every 10 minutes and contains 36792/5271/10540 samples for train/validation/test splits.
\end{itemize}

\end{document}

%% file: math_commands.tex
\usepackage{amsmath,amsfonts,bm}

\def\eqref#1{equation~\ref{#1}}

\def\1{\bm{1}}

\DeclareMathAlphabet{\mathsfit}{\encodingdefault}{\sfdefault}{m}{sl}
\SetMathAlphabet{\mathsfit}{bold}{\encodingdefault}{\sfdefault}{bx}{n}

